\documentclass[10pt,twocolumn,letterpaper]{article}

\usepackage{iccv}
\usepackage{authblk}
\usepackage{times}
\usepackage{epsfig}
\usepackage{graphicx}
\usepackage{amsmath}
\usepackage{amssymb}
\usepackage{enumitem}
\usepackage{multirow}
\usepackage{booktabs}
\usepackage{bbm}
\usepackage[T1]{fontenc}
\usepackage[utf8]{inputenc}
\usepackage[english]{babel}
\usepackage[autostyle, english=american]{csquotes}
\usepackage{subcaption}
\usepackage{ulem}

\MakeOuterQuote{"}
\usepackage{bbm}

\usepackage[breaklinks=true,bookmarks=false]{hyperref}

\iccvfinalcopy 

\def\iccvPaperID{****} 

\ificcvfinal\fi
\ificcvfinal\fi
\begin{document}

\title{Solution for SMART-101 Challenge of ICCV Multi-modal Algorithmic Reasoning Task 2023}


\author{
Xiangyu Wu$^{12}$,
Yang Yang$^1$,
Shengdong Xu$^1$,
Yifeng Wu$^2$,
Qingguo Chen$^2$,
Jianfeng Lu$^1$,
}

\affil{
 $^1$Nanjing University of Science and Technology
 
 $^2$Alibaba International Digital Commerce Group
}
\maketitle
\setlength{\intextsep}{1pt}
\setlength{\abovecaptionskip}{1.5pt}

\begin{abstract}
In this paper, we present our solution to a Multi-modal Algorithmic Reasoning Task: SMART-101 Challenge. Different from the traditional visual question answering datasets, this challenge evaluates the abstraction, deduction, and generalization abilities of neural networks in solving visuolinguistic puzzles designed specifically for children in the 6–8 age group. We employed a divide-and-conquer approach. At the data level, inspired by the challenge paper, we categorized the whole questions into eight types and utilized the llama-2-chat model to directly generate the type for each question in a zero-shot manner. Additionally, we trained a yolov7 model on the icon45 dataset for object detection and combined it with the OCR method to recognize and locate objects and text within the images. At the model level, we utilized the BLIP-2 model and added eight adapters to the image encoder VIT-G to adaptively extract visual features for different question types. We fed the pre-constructed question templates as input and generated answers using the flan-t5-xxl decoder. Under the puzzle splits configuration, we achieved an accuracy score of 26.5 on the validation set and 24.30 on the private test set.

\end{abstract}

\section{Introduction}
Visual question answering (VQA)~\cite{Salemi,relvit} task requires the model to take both the question Q in natural language and the image I as input and generate the answer A according to the information contained in the inputs. With the development of multi-modal large language model~\cite{blip2} technology, these models have demonstrated significant effectiveness in answering questions that require complex logical reasoning abilities.

\begin{figure}[ht]
    \centering
    \begin{subfigure}[t]{1.0\linewidth}
        \includegraphics[width=\textwidth]{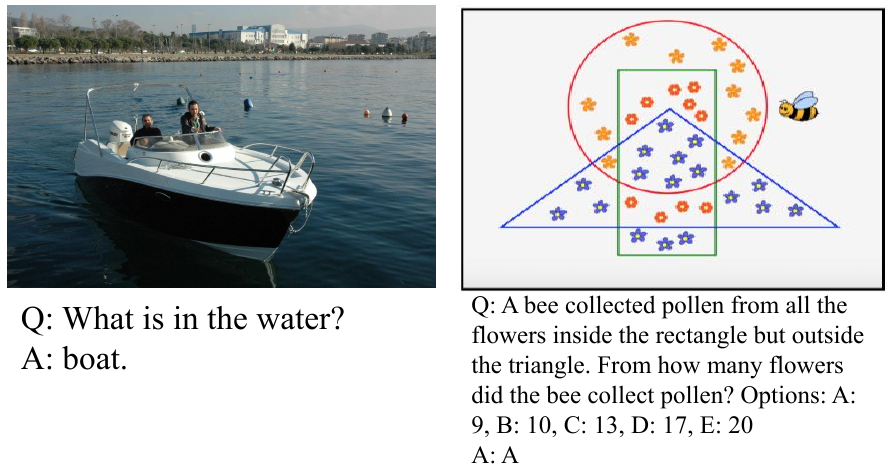}  
    \end{subfigure}
    \caption{The difference between traditional VQA and SMART-101 challenge.}
    \vspace{10pt}
    \label{fig: difference}
\end{figure}

Traditional visual question answering tasks primarily focus on real-world scene datasets, which evaluate the deep model's ability to recognize and locate objects in images and questions. By performing simple feature fusion, the model can provide accurate answers. As shown in Figure \ref{fig: difference}, the difficulty in this challenge is to evaluate the generalization abilities of deep neural networks in solving visuolinguistic puzzles designed specifically for children in the 6-8 age group and to understand the algorithmic reasoning abilities of SOTA deep models. This SMART-101~\cite{smart} dataset consists of 101 unique puzzles that require a mix of several elementary skills, including arithmetic, algebra, and spatial reasoning, among others. The currently deep models offer reasonable performances on puzzles in a supervised setting, they are not better than random accuracy when analyzed for generalization, and fail entirely on out-of-distribution generalization when the training and testing sets are disjoint at the puzzle levels. 

\vspace{10pt}

To address these challenges, we propose a divide-and-conquer approach. Firstly, we noticed the remarkable capability of large language models in zero-shot settings. Therefore, inspired by the SMART-101 challenge paper, we categorized the questions into eight types and used the llama-2-chat~\cite{llama2} model to directly generate the question type by constructing a proper prompt template in a zero-shot manner. Secondly, we observed that the object icons in the competition dataset are sourced from the icon45 dataset. Therefore, we automatically constructed an object detection dataset by randomly adding icons on a whiteboard background. Then, we trained a yolov7~\cite{yolov7} object detector and combined it with OCR~\cite{ocr} methods to locate and recognize objects and text in the images. Lastly, we employed the BLIP-2-flan-t5-xxl~\cite{blip2} model, which is a multi-modal large language model with strong capabilities in visual understanding and text generation. We added eight adapters to the image encoder VIT-G~\cite{vit} to adaptively extract visual features for different question types. We combined this information with candidate options to construct a question template, which was then passed as input to the BLIP-2 model to generate the final answer.

We introduce a divide-and-conquer approach, and its contributions can be summarized as follows:
\begin{itemize}
[itemsep=0pt,parsep=0pt,topsep=0pt,partopsep=0pt,leftmargin=*]
    \setlength{\itemindent}{1.3em}
    \item We propose the divide-and-conquer approach, utilizing large language models in a zero-shot paradigm to directly predict the type of each question.
    \item We automatically created an object detection dataset and trained an object detector and OCR model to locate and recognize objects and text in the images.
    \item We adopted the BLIP-2 multi-modal large language model and added multiple adapters to extract different visual features. Leveraging the powerful visual understanding and text generation capabilities of BLIP-2 to predict the answers.
\end{itemize}

\begin{figure*}
    \centering
    \includegraphics[scale=0.94]{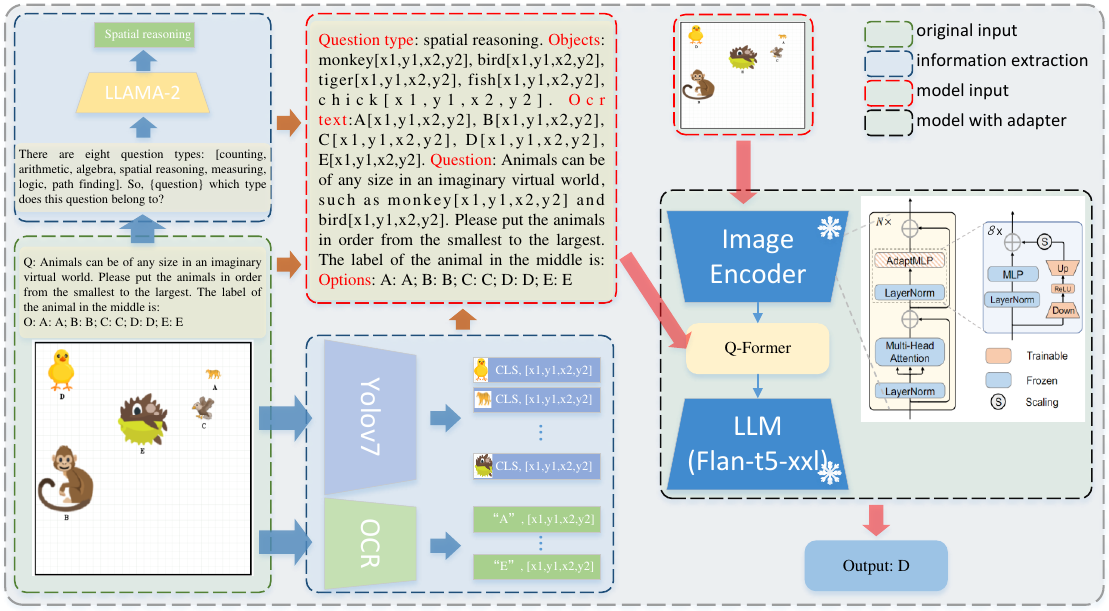}
    \caption{In the framework. "Original input" represents the raw image and question. "Information extraction" represents performing object detection, OCR, and generating the question type on the raw image and question. "Model input" consists of a question template and an image. "Model with adapter" refers to the addition of trainable adapters on the image encoder VIT-G of BLIP-2.}
    \label{fig: ourframework}
    \vspace{-10pt}
\end{figure*}

\section{Related Work}
\subsection{Large Language Model}
Large Language Models (LLMs) are AI models that can understand and generate human-like text. They are trained on a vast amount of text data and have been used in a variety of applications, such as translation, summarization, and coding. Some notable examples of LLMs include Chatgpt~\cite{glm} by OpenAI and flan-t5~\cite{flan-t5} by Google. However, these models also pose challenges, including potential misuse and the difficulty of controlling their output. Despite these challenges, the field of LLMs is advancing rapidly, with ongoing research aimed at improving their capabilities and addressing their limitations.

\subsection{Vision Adapter}
Vision Adapters~\cite{adaptformer,vit-adapter} are a recent development in the field of AI that aims to enhance the capabilities of existing models. For instance, the Vision Transformer Adapter is the first multi-task vision transformer adapter that learns generalizable task affinities which can be applied to novel tasks and domains. Another research is the CLIP-Adapter~\cite{vit-adapter} proposed by Peng Gao et al., which conducts fine-tuning with feature adapters on either visual or language branches. This approach has shown significant progress in visual representation learning. These adapters are integrated into an off-the-shelf vision transformer backbone and can simultaneously solve multiple dense vision tasks in a parameter-efficient manner. They outperform not only the existing convolutional neural network-based multitasking methods but also the vision transformer-based ones.

\subsection{Multi-modal Large Language Model}
In light of this complementarity, unimodal LLMs and vision models run towards each other at the same time, ultimately leading to the new field of Multi-modal Large Language Model (MLLM)~\cite{blip2,ofa,}. Formally, it refers to the LLM-based model with the ability to receive and reason with multi-modal information. It is able to receive and reason with information from multiple modalities, including text, images, and speech. MLLMs offer several advantages over unimodal LLMs, including more human-like perception, a more user-friendly interface, and the ability to support a larger range of tasks. MLLMs are seen as a potential step forward in the development of Artificial General Intelligence.

\section{Method}
The overall architecture of our approach is illustrated in Figure \ref{fig: ourframework}, consisting of four components: original input, information extraction, model input, and model with adapter.

\subsection{Original input}
The dataset for this challenge consists of an abstract image and a text input. The image is not a representation of a natural scene but rather an abstract composition of elements such as numbers, lines, boxes, and icons. The text input consists of a question and five options, with the questions being relatively complex and requiring strong logical reasoning abilities to generate correct answers.

\subsection{Information extraction}

Providing sufficient visual and textual information to the multi-modal model is crucial when dealing with complex multiple-choice visual question answering tasks. The dataset for this competition consists of 101 questions, categorized into eight types: counting, arithmetic, algebra, spatial reasoning, measuring, logic, and pathfinding. Each question type requires different visual information and reasoning abilities to answer. To better address these challenges, we employ a large language model using the zero-shot paradigm to predict the type of each question and reclassify the 101 questions accordingly. Specifically, we construct a prompt template as the input for the llama-2-chat~\cite{llama2} model, which states: "There are eight question types: [counting, arithmetic, algebra, spatial reasoning, measuring, logic, path finding]. So, {question} which type does this question belong to?". By utilizing this input format, the model can classify the questions based on their content and context. Considering the uncertainty in the large language model's outputs. For each question, we randomly selected 100 samples from all 2000 samples for a certain question for prediction. By aggregating the most frequently occurring output from multiple predictions, we determine the type of question. Through this method, we gain a better understanding of question types and provide more precise guidance for answering questions. This is crucial for successfully completing multiple-choice visual question answering tasks as each question type requires different reasoning and problem-solving approaches.

For this competition dataset, we noticed that it contains various elements in the images, such as numbers, English letters, animals, plants, and everyday objects. The presence of these elements inspired us to utilize object detection and OCR (Optical Character Recognition)~\cite{ocr} techniques for locating and recognizing text and icons in the images. To achieve this goal, we trained an object detector using the icon45 dataset. Specifically, we randomly selected n icons, each with a randomly assigned size, and placed them randomly on a whiteboard. This process allowed us to create a simple dataset for icon object detection. Next, we loaded the pre-trained weights of YOLOv7~\cite{yolov7} and trained an object detector capable of recognizing and localizing various icons in the images. This object detector helps us find and identify different icons in the images. For extracting textual information from the images, we employed the PaddleOCR~\cite{ocr} model for text recognition. This model assists in recognizing the textual content within the images and determining its position in the image. By combining object detection and OCR techniques, we can effectively extract important information from the images and utilize it for next processing and analysis.

\subsection{Model input}
Through the second step, we obtained the type of each question, as well as the category, position of the icons, and text content on the image. Then, we constructed a text template and input it into the BLIP-2~\cite{blip2} model. This template includes the type of question, all objects on the image (including their categories and positions), and the text content and position recognized by OCR. For example, as shown in Figure 2, for a spatial reasoning type question, we construct the following template: "Question type: spatial reasoning. Objects: monkey[x1,y1,x2,y2], bird[x1,y1,x2,y2], tiger[x1,y1,x2,y2], fish[x1,y1,x2,y2], chick[x1,y1,x2,y2]. Ocr text:A[x1,y1,x2,y2], B[x1,y1,x2,y2], C[x1,y1,x2,y2], D[x1,y1,x2,y2], E[x1,y1,x2,y2]. Question: Animals can be of any size in an imaginary virtual world, such as monkey[x1,y1,x2,y2] and bird[x1,y1,x2,y2]. Please put the animals in order from the smallest to the largest. The label of the animal in the middle is: Options: A: A; B: B; C: C; D: D; E: E." This template includes the type of question (spatial reasoning), all objects on the image (monkey, bird, tiger, fish, and chick) and their positions, text content (A, B, C, D, and E) recognized by OCR and their positions, as well as the specific description and options of the question. Then, we input this template together with the corresponding image into the BLIP-2 model for answer prediction. This method combines rich visual information and text information to effectively handle complex multiple-choice visual question answering tasks.

\subsection{Model with adapter}
In this competition task, we found that different types of questions require the extraction of different visual features for prediction. However, if a separate model is trained for each type of question, the cost would be very high. Therefore, we adopted a more efficient method. Specifically, we adopted the idea of adapters and added 8 visual adapters~\cite{adaptformer} to the image encoder VIT-G~\cite{vit} of the blip-2~\cite{blip2} model. Each adapter corresponds to a certain type of question and can adaptively extract the visual information needed for that type of question. In this way, we can handle various types of questions under a unified framework without having to train a separate model for each type of question. This method can not only greatly reduce the training cost but also improve the generalization ability of the model. Because each adapter is trained on the same model, they can share some parameters and structures of the model, thereby improving the generalization ability of the model.

\section{Experiment}
\textbf{Dataset.} The competition dataset is provided by the official organizers, which includes both a training set and a test set. The training data consists of 101 unique puzzles, with each puzzle having 2000 instances. Each puzzle sample contains an image, a question, and corresponding options. For each type of split, we will evaluate submissions on 100 puzzles. Furthermore, the new puzzle instances may originate from root puzzles that have never been encountered before.

\textbf{Metric.} In order to evaluate the model's exceptional generalization capability on SMART, we employ the segmentation method outlined in the original paper, referred to as Puzzle Split. This entails assessing novel, previously unobserved root puzzle instances that demand the same foundational skills as those in the training set. Recognizing the diverse difficulty levels of the puzzles, we also incorporate weights for each puzzle. Our chosen evaluation metric is the Weighted Option Selection Accuracy (WOSA), which is quantified by the following formula:

\begin{align}
    100 \times \frac{\sum_{i=1}^N w_i acc_i}{\sum_{i=1}^N w_i} 
\end{align}

where $w_i$ represents the weight of each puzzle in the test set, where $acc_i$ is 1 if the answer is correct, and 0 otherwise.

\textbf{Implementation Detail.} In our study, we trained our method based on BLIP-2 model. The training was conducted using 8 A100 GPUs, and the optimal performance was achieved at epoch 5. The learning rate was set to 3e-5, with a batch size of 32 per GPU.

\textbf{Result on private test set.} Table \ref{tab: res} shows the performance of our method on the private test set. From the results, compared to the baseline model in the original paper, our method has improved by 5.61\% on the accuracy (acc) metric. Due to the powerful text reasoning ability of the large language model, it has increased by 7.69\% on the \text{text\_wosa} metric. In addition, there are also significant improvements in the \text{vl\_wosa} and total \text{tot\_wosa} metrics.

\begin{table}
\centering
\begin{tabular}{ccccc}
\toprule
Method & acc & text\_wosa & vl\_wosa & tot\_wosa  \\
\hline
random &  0.93 &  0.00 &  2.72 &  1.59\\
baseline &  18.69 &  16.35 &  21.09 &  19.12\\
our &  24.30 &  24.04 &  21.77 &  22.71\\
\toprule
\end{tabular}
\caption{Private test set result.}
\label{tab: res}
\end{table}

\textbf{Ablation Study on evaluation set.} To analyze the contribution of each component of our method, we conduct more ablation studies on the evaluation set of the competition. From Table \ref{tab: abl}, compared to the blip-2-flan-t5-xxl model, the visual adapter has made a significant improvement. This is due to the extraction of different visual information for different types of questions. In addition, the text information on the image and the category and position of icons of the image also provides more abundant visual information for the model.

\begin{table}[htp]
    \centering
    \begin{tabular}{cccccc}
    \toprule
    Method & ACC \\
    \hline
    BLIP-2-Flan-t5-xxl & 21.1 \\
    \hspace{3ex}+ adapter & 24.3 \\
    \hspace{3ex}+ OCR & 25.1 \\
    \hspace{3ex}+ YOLOv7 & 26.5 \\
    \toprule
    \end{tabular}
\caption{Ablation experiment.} 
\label{tab: abl}
\end{table}

\section{Conclusion}

In this competition, we found that providing sufficient input information to the model is crucial. This includes the type of question, as well as the category, position, and text content of the icons on the image. We used a large language model to predict the type of each question through a zero-shot paradigm and constructed a text template to input into the model. In addition, we also used object detection and OCR technology to locate and recognize the text and icons on the image. Finally, we input all this information into the model for answer prediction. This method, which combines rich visual information and text information, can effectively handle complex multiple-choice visual question answering tasks and has achieved good results.

{\small
\bibliographystyle{ieee_fullname}
\bibliography{mybib}
}
\end{document}